\documentclass{article}

\usepackage{arxiv}

\usepackage[utf8]{inputenc} 
\usepackage[T1]{fontenc}    
\usepackage{hyperref}       
\usepackage{url}            
\usepackage{booktabs}       
\usepackage{amsfonts}       
\usepackage{nicefrac}       
\usepackage{microtype}      
\usepackage{lipsum}
\usepackage{graphicx}
\usepackage{caption}
\usepackage{subcaption}
\usepackage{xcolor}
\graphicspath{ {./images/} }

\title{Multi-pooled Inception features for no-reference image quality assessment}

\author{
 Domonkos Varga \\
}

\begin{document}
\maketitle
\begin{abstract}
Image quality assessment (IQA)
is an important element of a broad spectrum of applications
ranging from automatic video streaming to display technology.
Furthermore, the measurement
of image quality requires a balanced investigation of image
content and features.
Our proposed approach extracts visual features by attaching
global average pooling (GAP) layers to multiple Inception modules
of on an ImageNet database pretrained convolutional neural network (CNN).
In contrast to previous methods, we do not
take patches from the input image. Instead, the
input image is treated as a whole and is run through a pretrained CNN body
to extract resolution-independent, multi-level deep features.
As a consequence, our method can be easily generalized to
any input image size and pretrained CNNs. Thus,
we present a detailed parameter study
with respect to the CNN base architectures and the effectiveness of different
deep features. We demonstrate that our best
proposal --- called MultiGAP-NRIQA ---
is able to provide state-of-the-art results on three benchmark IQA databases.
Furthermore, these results were also confirmed in a cross database test
using the LIVE In the Wild Image Quality Challenge database.
\end{abstract}


\section{Introduction}
\label{sec:intro}
With the increasing popularity of imaging devices as
well as the rapid spread of social media and multimedia sharing
websites, digital images and videos have become an essential part
of daily life, especially in everyday communication.
Consequently, there is a growing need for effective systems
that are able to monitor the quality of visual signals.
Obviously, the most reliable way of assessing image quality
is to perform subjective user studies, which involves the
gathering of individual quality scores.
However, the compilation and evaluation of a subjective
user study are very slow and
laborious processes.
Furthermore, their application in a real-time system is impossible.
In contrast, objective image quality assessment (IQA)
involves the development of quantitative measures and algorithms
for estimating image quality.
    
Objective IQA is classified based on the availability of the reference image.
Full-reference image quality assessment (FR-IQA) methods have full
access to the reference image, whereas
no-reference image quality assessment (NR-IQA) algorithms
possess only the distorted digital image.
In contrast, reduced-reference image quality assessment (RR-IQA)
methods have partial information about the
reference image;
for example, as a set of extracted features.
Objective IQA algorithms are evaluated
on benchmark databases containing the distorted 
images and their corresponding mean opinion scores (MOSs),
which were collected during subjective user studies.
The MOS is a real number, typically in the range 1.0--5.0, where
1.0 represents
the lowest quality and 5.0 denotes the best quality.
Furthermore, the MOS of an image is its arithmetic mean over all collected
individual quality ratings.
As already mentioned,
publicly available IQA databases help researchers to devise and
evaluate IQA algorithms and metrics. Existing IQA datasets can be
grouped into two categories with respect to the introduced
image distortion types. The first category contains images
with artificial distortions, while the images of the second category
are taken from sources with "natural" degradation without any
additional artificial distortions.

The rest of this section is organized as follows. In Subsection
\ref{sec:related}, we review related work in NR-IQA with a special
attention on deep learning based methods. Subsection
\ref{sec:contributions} introduces the contributions made in this study.
\subsection{Related work}
\label{sec:related}
Many traditional NR-IQA algorithms rely on the so-called natural scene statistics (NSS) \cite{reinagel1999natural} model.     
These methods assume that natural images possess a
particular regularity that is modified by visual
distortion. Further, by quantifying the deviation
from the natural statistics, perceptual image quality can be
determined. NSS-based feature vectors usually rely 
on the wavelet transform \cite{moorthy2011blind}, discrete cosine
transform \cite{saad2011dct}, curvelet transform \cite{liu2014no},
shearlet transform \cite{li2014no},
or transforms to other
spatial domains \cite{mittal2012no}.
DIIVINE \cite{moorthy2011blind} (Distortion Identification-based Image Verity and INtegrity Evaluation) exploits
NSS using wavelet transform and consists of two steps.
Namely, a probabilistic distortion identification stage is followed
by a distortion-specific quality assessment one.
In contrast, He \textit{et al.} \cite{he2012sparse} presented a sparse feature
representation of NSS using also the wavelet transform. 
Saad \textit{et al.} \cite{saad2011dct} built a feature vector
from DCT coefficients. Subsequently, a Bayesian inference approach
was applied for the prediction of perceptual quality scores.
In \cite{garcia2018application}, the authors presented a detailed
review about the use of local binary pattern texture descriptors
in NR-IQA.

Another line of work focuses on opinion-unaware algorithms
that require neither training samples nor human
subjective scores.
Zhang \textit{et al.} \cite{zhang2015feature} introduced 
the integrated local natural image quality evaluator (IL-NIQE), which
combines features of NSS with multivariate Gaussian models of image patches.
This evaluator uses several quality-aware
NSS features, i.e., the statistics of normalized luminance, 
mean subtracted and contrast-normalized products of pairs
of adjacent coefficients,
gradient, log-Gabor filter responses, and color (after 
the transformation into a logarithmic-scale opponent color space).

Kim \textit{et al.} \cite{kim2016fully} introduced a
no-reference image quality predictor called the blind image evaluator based on
a convolutional neural network (BIECON), in which the
training process is carried out in two steps. First, local metric score regression and then
subjective score regression are conducted. During the local metric score
regression, non-overlapping image patches are trained independently;
FR-IQA methods such as SSIM or GMS are used for the target patches.
Then, the CNN trained on image patches is refined by targeting the
subjective image score of the complete image. Similarly,
the training of a multi-task end-to-end optimized deep neural
network \cite{ma2017end} is carried out in two steps.
Namely, this architecture contains two sub-networks:
a distortion identification network and a quality prediction network.
Furthermore, a biologically inspired generalized divisive
normalization \cite{li2009reduced} is applied as
the activation function in the network instead of rectified linear units (ReLUs).
Similarly, Fan \textit{et al.} \cite{fan2018no} introduced a two-stage framework.
First, a distortion type classifier identifies the distortion type then
a fusion algorithm is applied to aggregate the results of expert networks
and produce a perceptual quality score.

In recent years, many algorithms relying on deep learning have been proposed. Because of the small size of many existing image quality benchmark
databases, most deep learning based methods employ CNNs as feature extractors or take patches from the training images to increase the
database size. The CNN framework of Kang \textit{et al.} \cite{kang2014convolutional} is trained on non-overlapping
image patches extracted from
the training images.
Furthermore, these patches inherit the MOS of their source images.
For preprocessing, local contrast normalization is employed.
The applied CNN consists of conventional building
blocks, such as convolutional, pooling, and fully connected layers.
Bosse \textit{et al.} \cite{bosse2016deep} introduced a similar method.
Namely, they developed a 12-layer CNN that is trained
on $32\times32$ image patches.
Furthermore, a weighted average patch aggregation method
was introduced in which weights representing the relative
importance of image patches in quality
assessment are learned by a subnetwork.
In contrast, Li \textit{et al.} \cite{li2016no} combined a
CNN trained on image patches with the Prewitt magnitudes of
segmented images to predict perceptual quality.

Li \textit{et al.} \cite{li2017no} trained a CNN on $32\times 32$
image patches and employed it as a feature extractor.
In this method, a feature vector of length $800$ represents each
image patch of an input image and the sum of image patches' feature vectors is
associated with the original input image. Finally, a support
vector regressor (SVR) is trained to evaluate the image quality using
the feature vector representing the input image.
In contrast, Bianco \textit{et al.} \cite{bianco2018use} utilized
a fine-tuned AlexNet \cite{alexnet}
as a feature extractor on the target database.
Specifically, image quality is predicted by averaging
the quality ratings on multiple randomly
sampled image patches. Further, the perceptual quality
of each patch is predicted by an SVR trained
on deep features extracted with the help of a fine-tuned AlexNet \cite{alexnet}.
Similarly, Gao \textit{et al.} \cite{gao2018blind} employed
a pretrained CNN as a feature extractor, but they generate
one feature vector for each CNN layer.
Furthermore, a quality score is predicted for each feature
vector using an SVR. Finally, the overall perceptual quality of the image is
determined by averaging these quality scores.
In contrast, Zhang \textit{et al.} \cite{zhang2018blind}
trained first a CNN to identify image
distortion types and levels.
Furthermore, the authors took another CNN, that was trained on ImageNet,
to deal with authentic distortions. To predict perceptual image quality,
the features of the last convolutional layers were pooled bi-linearly and
mapped onto perceptual quality scores with a fully-connected layer.
He \textit{et al.} \cite{he2019visual} proposed a method
containing two steps. In the
first step, a sequence of image patches is created from the input image.
Subsequently, features are extracted with the help of a CNN and
a long short-term memory (LSTM) is utilized to evaluate the level of image
distortion. In the second stage, the model is trained to predict the
patches' quality score. Finally, a saliency weighted procedure is
applied to determine the whole image's quality from the patch-wise
scores. Similarly, Ji \textit{et al.} \cite{ji2019blind} utilized a CNN
and an LSTM for NR-IQA, but the deep features were extracted from the
convolutional layers of a VGG16 \cite{simonyan2014very} network.
In contrast to other algorithms,
Zhang \textit{et al.} \cite{zhang2018deep}
proposed an opinion-unaware deep method. Namely, high-contrast image patches
were selected using deep convolutional maps 
from pristine images which were used to train a multi-variate
Gaussian model.

\subsection{Contributions}
\label{sec:contributions}
Convolutional neural networks (CNNs) have demonstrated great
success in a wide range of
computer vision
tasks \cite{varga2019no}, \cite{iandola2016squeezenet}, \cite{gordo2016deep},
including
NR-IQA \cite{kang2014convolutional}, \cite{bosse2016deep}, \cite{li2016no},
\cite{alaql2019no}.
Furthermore, pretrained CNNs can also provide a useful feature
representation for a variety of tasks \cite{sharif2014cnn}.
In contrast, employing pretrained CNNs is not straightforward.
One major challenge is that CNNs require a fixed input size.
To overcome this constraint,
previous methods for
NR-IQA \cite{kang2014convolutional}, \cite{bosse2016deep}, \cite{li2016no}, \cite{bianco2018use}
take patches from the input image. 
Furthermore, the evaluation of perceptual
quality was based on these image patches or on
the features extracted from them.
In this paper, we make the following contributions.
We introduce a unified and content-preserving architecture
that relies on the Inception modules of 
pretrained CNNs, such as GoogLeNet \cite{szegedy2015going} or
Inception-V3 \cite{szegedy2016rethinking}.
Specifically, this novel architecture applies
visual features extracted from multiple Inception modules
of pretrained CNNs and
pooled by
global average pooling (GAP) layers.
In this manner, we obtain both intermediate-level and high-level
representation from CNNs and each level of representation is considered
to predict image quality.
Due to this architecture, we do not take patches from the input image
like previous methods \cite{kang2014convolutional},
\cite{bosse2016deep},
\cite{li2016no}, \cite{bianco2018use}.
Unlike previous deep architectures \cite{he2019visual}, \cite{bosse2016deep},
\cite{bianco2018use}
we do not utilize only the deep
features of the last layer of a pretrained CNN.
Instead, we carefully examine the effect of different features
extracted from different layers on the prediction performance and we point
out that the combination of deep features from mid- and high-level
layers results in significant prediction performance increase.
With experiments on three publicly available
benchmark databases, we demonstrate
that the proposed method is able to outperform other
state-of-the-art methods.
Specifically, we
utilized KonIQ-10k \cite{lin2018koniq}, KADID-10k \cite{lin2019kadid}, and
LIVE In the Wild Image Quality Challenge Database \cite{ghadiyaram2015massive}
databases.
KonIQ-10k \cite{lin2018koniq} is the largest publicly available
database containing 10,073 images with
authentic distortions,
while KADID-10k \cite{lin2019kadid} consists of 81 reference images and 10,125
distorted ones (81 reference images $\times$ 25 types of
distortions $\times$ 5 levels of distortions). 
LIVE In the Wild Image Quality Challenge Database \cite{ghadiyaram2015massive} is
significantly smaller than KonIQ-10k \cite{lin2018koniq} or
KADID-10k \cite{lin2019kadid}.
For a cross database test,
also the LIVE In the Wild Image Quality Challenge Database \cite{ghadiyaram2015massive}
is applied
which contains $1,162$ images with authentic distortions evaluated by over $8,100$
unique human observers.

The remainder of this paper is organized as follows. After this introduction, Section 2 introduces our proposed approach.
In Section 3, the experimental results and analysis are
presented, and a conclusion is drawn in Section 4.

\section{Methodology}
\label{sec:methods}
\begin{figure}[!ht]
\centering
\includegraphics[width=4.0in]{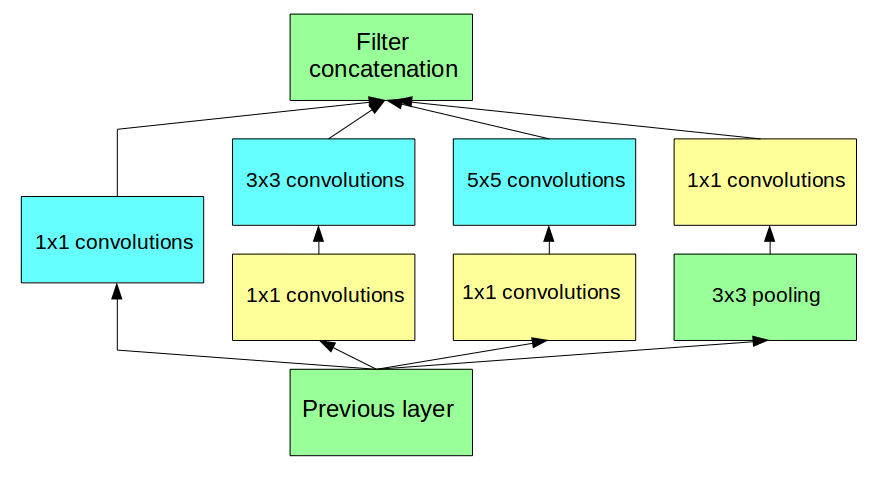}
\caption{Illustration of Inception module.
It was restricted to filter sizes
$1\times1$, $3\times3$, and $5\times5$.
Subsequently, the outputs were concatenated into a single vector
that is the input for the next stage.
Adding of an alternative parallel pooling path was found to be
beneficial. Applying filters
of $1\times1$ convolution makes possible to reduce the volume
before the expensive $3\times3$ and $5\times5$
convolutions \cite{szegedy2015going}.
}
\label{inception}
\end{figure}

To extract visual features, GoogLeNet \cite{szegedy2015going} or
Inception-V3 \cite{szegedy2016rethinking}
were applied
as base models. 
GoogLeNet \cite{szegedy2015going} is a 22 layer deep
CNN and was the winner of ILSVRC 2014 with
a top 5 error rate of 6.7 \%.
Depth and width of the network was increased but not
simply following the general method of stacking
the layers on each other. A new
level of organization was introduced
codenamed Inception module (see Figure \ref{inception}).
In GoogLeNet \cite{szegedy2015going} not
everything happens sequentially like in previous
CNN models, pieces of the network work in parallel.
Inspired by a neuroscience model in \cite{serre2007robust}
where for handling multiple scales a series of Gabor
filters were used with a two layer deep model.
But contrary to the beforementioned model all 
layers are learned and not fixed.
In GoogLeNet \cite{szegedy2015going} architecture Inception
layers are introduced and repeated many times.
Subsequent improvements of GoogLeNet \cite{szegedy2015going} have been called
Inception-v$N$ where $N$ refers to the version number put out by Google.
Inception-V2 \cite{szegedy2016rethinking} 
was refined by the introduction of
batch normalization \cite{ioffe2015batch}.
Inception-V3 \cite{szegedy2016rethinking} was improved by
factorization ideas. Factorization into smaller
convolutions means for example replacing a $5\times5$ convolution by
a multi-layer network
with fewer parameters but with the same input size and output depth.

We chose the features of Inception modules for the following reasons.
The main motivation behind the construction of Inception modules is that
salient parts of images may very extremely.
This means that the region of interest can occupy
very different image regions both in terms of size and location.
That is why, determining the convolutional kernel size in a CNN is very
difficult. Namely, a larger kernel size is required for visual information
that is distributed rather globally. On the other hand, a smaller kernel size
is better for visual information that is distributed more locally.
As already mentioned, the creators of Inception modules reflected to
this challenge by the introduction of multiple filters with
multiple sizes on the same level.
Furthermore, visual distortions have a similar nature. Namely, the distortion
distribution is strongly influenced by image content \cite{gu2016analysis}.

\begin{figure*}[!ht]
\centering\includegraphics[width=5.0in]{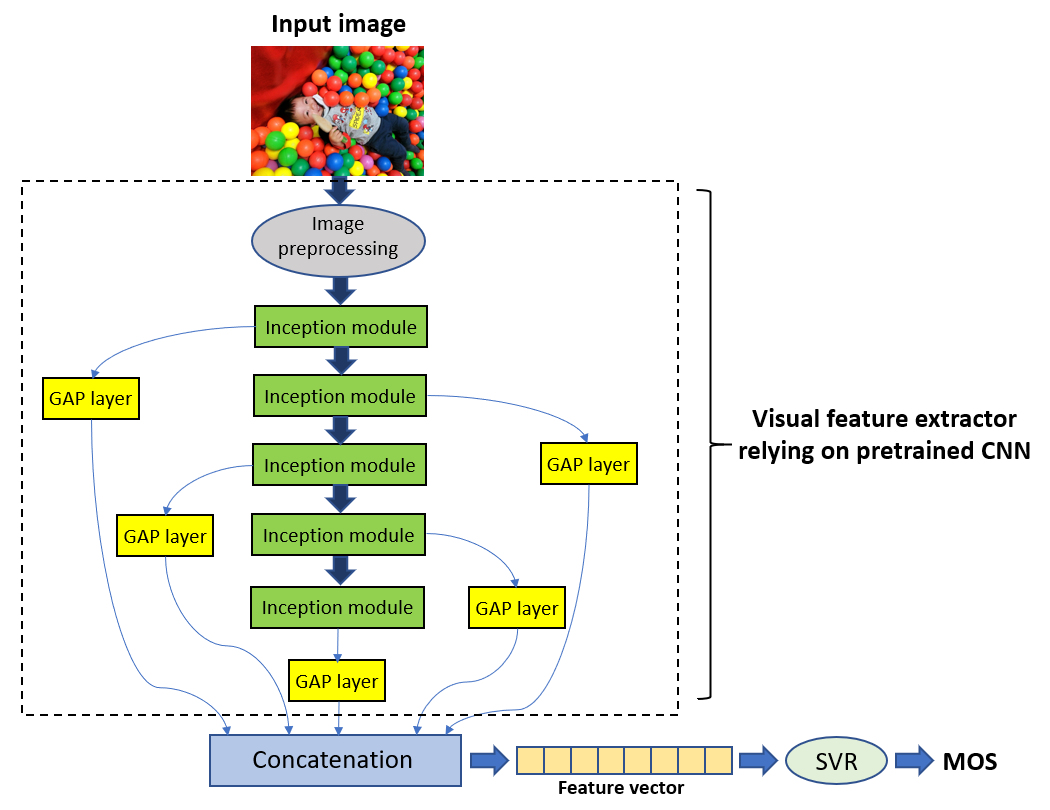}
\caption{The pipeline of the proposed method.
An input image is run through on an ImageNet
database pretrained
CNN body (GoogLeNet
and
Inception-V3 are considered 
in this study) which carries out all its
defined operations. Furthermore, global average pooling (GAP) layers
are attached to each Inception module to extract resolution independent
deep features at different abstraction levels. The feature vectors obtained
from the Inception modules are concatenated and an SVR with radial basis
function is applied to predict perceptual image quality.
}
\label{overview}
\end{figure*}


\subsection{Pipeline of the proposed method}
The pipeline of the proposed framework is depicted in Figure
\ref{overview}.
A given input image to be evaluated is run through a pretrained CNN
body (GoogLeNet \cite{szegedy2015going} and
Inception-V3 \cite{szegedy2016rethinking} are
considered in this study)
which carries out all its defined operations.
Specifically, global average pooling (GAP) layers are attached to the output
of each Inception module.
Similar to max- or min-pooling layers, GAP layers are applied in
CNNs to reduce the spatial dimensions of convolutional layers.
However, a GAP layer carries out a more extreme type of dimensional
reduction than a max- or min-pooling layer.
Namely, an $h\times w \times d $ block is reduced to $1\times 1 \times d $.
In other words, a GAP layer reduces 
a feature map to a single value by taking the average of
this feature map.
By adding GAP layers to each Inception module, we are able
to extract resolution independent features at different levels of
abstraction. Namely, the feature maps produced by 
neuroscience models inspired \cite{serre2007robust}
Inception modules have been shown representative for
object categories \cite{szegedy2015going}, \cite{szegedy2016rethinking}
and correlate well with human
perceptual quality judgments \cite{zhang2018unreasonable}.
The motivation behind the application of GAP layers was the followings.
By attaching GAP layers to the Inception modules, we gain an architecture which
can be easily generalized to any input image resolution and base CNN architecture.
Furthermore, this way the decomposition of the input image into smaller patches
can be avoided which means that parameter settings related to the
database properties (patch size, number of patches, sampling strategy, \textit{etc.})
can be ignored. Moreover, some kind of image distortions are not uniformly distributed
in the image. These kind of distortions could be better captured in an aspect-ratio
and content preserving architecture.

As already mentioned, a feature vector is extracted over each
Inception module using a GAP layer. Let $\textbf{f}_k$ denote the
feature vector extracted from the $k$th Inception module. The input image's
feature vector is obtained by concatenating the respective feature vectors
produced by the Inception modules. Formally, we can write
$\textbf{F}=\textbf{f}_1\oplus\textbf{f}_2\oplus...\oplus\textbf{f}_N$,
where $N$ denotes the number of Inception modules in the base CNN and
$\oplus$ stands for the concatenation operator. In Section
\ref{sec:param}, we present a detailed analysis about the effectiveness
of different Inception modules' deep features as a perceptual metric.
Furthermore, we point out the prediction performance increase due to the
concatenation of deep features extracted from different abstraction levels.

Subsequently, an SVR \cite{drucker1997support} with radial basis function (RBF) kernel
is trained to learn the mapping
between feature vectors and corresponding perceptual quality scores.

Moreover, we also applied Gaussian process regression (GPR) with rational
quadratic kernel function \cite{rasmussen2003gaussian} in Section \ref{sec:comp}.

\subsection{Database compilation and transfer learning}
Many image quality assessment databases are available online, such as
TID2013 \cite{ponomarenko2015image} or 
LIVE In the Wild \cite{ghadiyaram2015massive},
for research purposes. In this study, we
selected the recently published KonIQ-10k \cite{lin2018koniq}
database to train and test
our system, because it is the largest available database containing digital
images with authentic distortions. Furthermore, we present a parameter study
on KonIQ-10k \cite{lin2018koniq} to find the best design choices.
Our best proposal is compared
against the state-of-the-art on KonIQ-10k \cite{lin2018koniq} and
also on other publicly
available databases.

\begin{figure}[!ht]
\centering\includegraphics[width=4.0in]{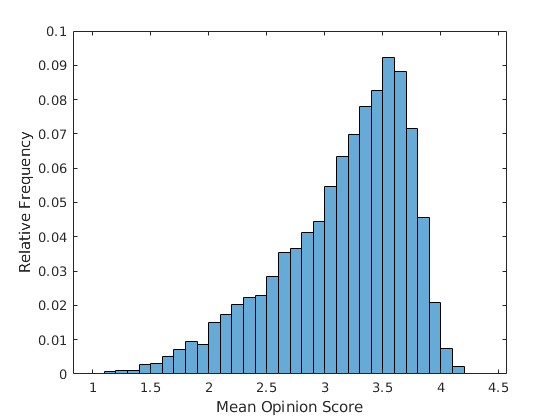}
\caption{MOS distribution in KonIQ-10k database.
It contains 10,073 RGB images with authentic distortions and the corresponding
MOS values which are on a scale from 1.0 (worst image quality) to
5.0 (best image quality).
}
\label{koniq}
\end{figure}


KonIQ-10k \cite{lin2018koniq} consists of 10,073 digital 
images with the corresponding MOS
values. To ensure the fairness of the experimental setup, we selected
randomly 6,073 images $(\sim60\%)$ for training,
2,000 images $(\sim20\%)$ for validation, and
2,000 images $(\sim20\%)$ for testing purposes.
First, the
base CNN was fine-tuned on target
database KonIQ-10k \cite{lin2018koniq} using the
above-mentioned training
and the validation subsets.
To this end, regularly the base CNN's last 1,000-way softmax
layer was removed and replaced
by a 5-way one in previous methods \cite{bianco2018use},
because the training and validation
subsets were reorganized into
five classes with respect to the MOS values: class $A$ for excellent
image quality $(5.0>MOS\geq4.2)$, class $B$ for good image quality
$(4.2>MOS\geq3.4)$, class $C$ for fair image quality $(3.4>MOS\geq2.6)$, class
$D$ for poor image quality $(2.6>MOS\geq1.8)$, and class $E$ for very
poor image quality $(1.8>MOS\geq1.0)$. Subsequently, the base CNN
was further train to classify the images into quality categories.
Since the MOS distribution in KonIQ-10k \cite{lin2018koniq} is
strongly imbalanced (see Figure \ref{koniq}), there would be
very little number of
images in the class for excellent images. That is why, we took
a regression-based approach instead of classification-based
approach for fine-tuning. Namely, we removed the base CNN's
last 1,000-way softmax
layer and we replaced it by a regression layer containing only one neuron.
Since GoogLeNet \cite{szegedy2015going} and
Inception-V3 \cite{szegedy2016rethinking} accept images with input size of
$224\times224$ and $299\times299$, respectively, \textit{twenty}
$224\times224$-sized or $299\times299$-sized patches were cropped
randomly from each training and validation images.
Furthermore, these patches inherit the perceptual
quality score of their source images and the fine-tuning
is carried out on these patches.
Specifically, we trained the base CNN further for regression to predict
the images patches MOS values which are inherited from their source images.
During fine-tuning Adam optimizer \cite{kingma2014adam} was
used, the initial
learning rate was set to 0.0001 and divided by 10
when the validation error stopped improving. Further,
the batch size was set to 28 and the momentum was 0.9 during
fine-tuning. 

\section{Experimental results and analysis}
\label{sec:exp}
In this section, we demonstrate our experimental results.
First, we give the definition of the evaluation metrics in Section 
\ref{sec:metrics}. Second, we describe the experimental setup
and the implementation details in
Section \ref{sec:exps}. In Section \ref{sec:param}, we give a
detailed parameter
study to find the best design choices of the proposed
method using KonIQ-10k \cite{lin2018koniq}
database.
Subsequently, we carry out a comparison to other state-of-the-art methods
using KonIQ-10k \cite{lin2018koniq}, KADID-10k \cite{lin2019kadid}, 
and LIVE In the Wild \cite{ghadiyaram2015massive}
publicly available IQA databases.
Finally, we present a so-called cross database test using
LIVE In the Wild Image Quality
Challenge database \cite{ghadiyaram2015massive}.
\subsection{Evaluation metrics}
\label{sec:metrics}
The performance of NR-IQA algorithms are characterized by the correlation
calculated between the ground-truth scores of a benchmark database and
the predicted scores. To this end, Pearson's linear correlation coefficient (PLCC)
and Spearman's rank order correlation coefficient (SROCC) are widely used
in the literature \cite{xu2015visual}. 
PLCC between datasets $A$ and $B$ is defined as
\begin{equation}
PLCC(A, B)=\frac{\sum_{i=1}^m (A_i-\bar{A})(B_i-\bar{B})}{\sqrt{\sum_{i=1}^m (A_i-\bar{A})^2}\sqrt{\sum_{i=1}^m (B_i-\bar{B})^2}},
\end{equation}
where $\bar{A}$ and $\bar{B}$ denote the average of
sets $A$ and $B$, and $A_i$ and $B_i$ denote the $i$th elements
of sets $A$ and $B$, respectively.
SROCC, it can be expressed as
\begin{equation}
SROCC(A, B)=\frac{\sum_{i=1}^m (A_i-\hat{A})(B_i-\hat{B})}{\sqrt{\sum_{i=1}^m (A_i-\hat{A})^2}\sqrt{\sum_{i=1}^m (B_i-\hat{B})^2}},    
\end{equation}
where $\hat{A}$ and $\hat{B}$ stand for the middle ranks.
\subsection{Experimental setup and implementation details}
\label{sec:exps}
As already mentioned, a detailed parameter study was carried out
on the recently published KonIQ-10k \cite{lin2018koniq},
which is the
currently largest available
IQA database with authentic distortions,
to determine the optimal
design choices. Subsequently, our best proposal is compared to the
state-of-the-art using other publicly available databases as well.

The proposed method was implemented in MATLAB R2019a mainly relying
on the functions of the Deep Learning Toolbox (formerly Neural
Network Toolbox), Image Processing Toolbox, and Statistics and
Machine Learning Toolbox.
Thus, the parameter study was also carried out in MATLAB environment.
More specifically, it was evaluated by 100 random
train-validation-test split of the applied database 
and we report on the average of the PLCC and SROCC values.
As usual in machine learning, $\sim60\%$ of the images
was used for training, $\sim20\%$ for validation, and $\sim20\%$
for testing purposes.
Moreover, for IQA databases containing artificial distortions the splitting
of the database is carried out with respect to the reference images, so
no semantic overlapping was between the training, validation, and test sets.
Further, the models were trained
and tested on a personal computer
with 8-core i7-7700K CPU 
two NVidia Geforce GTX 1080 GPUs. 
\subsection{Parameter study}
\label{sec:param}
First, we conducted experiments to determine which
Inception module in GoogLeNet \cite{szegedy2015going}
or in Inception-V3 \cite{szegedy2016rethinking} is the most
appropriate for visual feature extraction to
predict perceptual image quality. Second, we answer the question whether the
concatenation of different Inception modules' feature vectors improves the
prediction's performance or not. Third, we demonstrate that
fine-tuning of the base
CNN architecture results in significant performance increase.
In this parameter study, we used KonIQ-10k database to answer the
above mentioned questions and to find the most effective design choices.
In the next subsection, our best proposal is used to carry out a comparison
to the state-of-the-art using other databases as well.

The results of the parameter study are summarized in Tables
\ref{table:param1a}, \ref{table:param1b}, \ref{table:param2a}, and
\ref{table:param2b}. Specifically, Table \ref{table:param1a} and 
\ref{table:param2a} contains the results
with GoogLeNet \cite{szegedy2015going} and
Inception-V3 \cite{szegedy2016rethinking}
base architectures without fine-tuning, respectively. On the other hand,
Table \ref{table:param1b} and \ref{table:param2b} summarizes the results
when fine-tuning is applied. In these tables, we reported on the average,
the median, and the standard deviation of the PLCC and SROCC values obtained
after 100 random train-validation-test splits using KonIQ-10k database. 
Furthermore, we report
on the effectiveness of deep features extracted from different
Inception modules.
Moreover, the tables also contain the prediction performance of the
concatenated deep feature vector. From these results, it can be
concluded that the deep features extracted from the
early Inception modules perform
slightly poorer than those of intermediate and last Inception modules.
Although most state-of-the-art
methods \cite{he2019visual}, \cite{bosse2016deep}, \cite{bianco2018use}
utilize the features of the last CNN layers, it is
worth to examine earlier layers as well, because the
tables' data indicate that the middle layers
encode those information which are the most powerful for perceptual
quality prediction. We can also assert that feature vectors containing
both mid-level and high-level deep representations are significantly
more efficient than those of containing only one level's feature
representation. Finally, it can be clearly seen that fine-tuning the
base CNN architectures also improves the effectiveness of the extracted
deep features. On the whole, the deeper
Inception-V3 \cite{szegedy2016rethinking} provides more effective
features than GoogLeNet \cite{szegedy2015going}. Our best proposal relies
on Inception-V3 and concatenates the features of all Inception modules.
In the followings, we call this architecture 
\textit{MultiGAP-NRIQA} and compare it
to other state-of-the-art in the next subsection.

Another contribution of this parameter study may be the followings. It is
worth to study the features of different layers separately because the features
of intermediate layers may provide a better representation of the given task
than high-level features. Furthermore, the proposed feature extraction method
may be also superior in other problems where the task is to predict one value only from
the image data itself relying on a large enough database.

In our environment (MATLAB R2019a, PC with 8-core i7700K CPU and two
NVidia Geforce GTX 1080), the computational times of the proposed
MultiGAP-NRIQA method are the followings. The loading of the base CNN
and the $1024\times 768$-sized or the $512\times384$
input image takes about $1.8 s$. Furthermore, the
feature extraction from multiple Inception
modules of Inception-V3 \cite{szegedy2016rethinking}
and concatenation takes on average
$1.355 s$ or $0.976 s$ on the GPU, respectively.
Furthermore, the SVR regression takes
$2.976 s$ on average computing on the CPU.

\begin{table*}[ht]
\caption{Performance comparison of deep features extracted from
GoogLeNet's Inception modules without 
fine-tuning measured on KonIQ-10k. 
Average/median $(\pm std)$ values are
reported over 100 random train-test splits. The best results are typed
by \textbf{bold}.
} 
\centering 
\begin{center}
    \begin{tabular}{ |c|c|c|c|}
    \hline
Layer&Dimension&PLCC&SROCC\\
    \hline
\textit{inception\_3a-output}&256&$0.845/0.845 (\pm0.006)$&$0.842/0.841 (\pm0.007)$\\
\textit{inception\_3b-output}&480&$0.861/0.861 (\pm0.007)$&$0.856/0.858 (\pm0.007)$\\
\textit{inception\_4a-output}&512&$0.876/0.876 (\pm0.004)$&$0.872/0.872 (\pm0.006)$\\
\textit{inception\_4b-output}&512&$0.874/0.874 (\pm0.005)$&$0.865/0.864 (\pm0.008)$\\
\textit{inception\_4c-output}&512&$0.875/0.877 (\pm0.006)$&$0.865/0.865 (\pm0.006)$\\
\textit{inception\_4d-output}&528&$0.876/0.875 (\pm0.007)$&$0.864/0.864 (\pm0.007)$\\
\textit{inception\_4e-output}&832&$0.872/0.871 (\pm0.006)$&$0.861/0.862 (\pm0.005)$\\
\textit{inception\_5a-output}&832&$0.873/0.874 (\pm0.005)$&$0.859/0.860 (\pm0.005)$\\
\textit{inception\_5b-output}&1024&$0.861/0.861 (\pm0.008)$&$0.851/0.850 (\pm0.008)$\\
 \hline
\textit{All concatenated}&5488&$\textbf{0.889/0.889} (\pm0.007)$&$\textbf{0.879/0.877} (\pm0.006)$\\ 
\hline
 \end{tabular}
\end{center}
\label{table:param1a}
\end{table*}

\begin{table*}[ht]
\caption{Performance comparison of deep features extracted from
GoogLeNet's Inception modules with
fine-tuning measured on KonIQ-10k. 
Average/median $(\pm std)$ values are
reported over 100 random train-validation-test
splits. The best results are typed
by \textbf{bold}.
} 
\centering 
\begin{center}
    \begin{tabular}{ |c|c|c|c|}
    \hline
Layer&Dimension&PLCC&SROCC\\
    \hline
\textit{inception\_3a-output}&256&$0.850/0.849 (\pm0.007)$&$0.846/0.846 (\pm0.007)$\\
\textit{inception\_3b-output}&480&$0.866/0.866 (\pm0.006)$&$0.861/0.862 (\pm0.007)$\\
\textit{inception\_4a-output}&512&$0.881/0.881 (\pm0.005)$&$0.877/0.876 (\pm0.006)$ \\
\textit{inception\_4b-output}&512&$0.877/0.876 (\pm0.005)$&$0.870/0.870 (\pm0.006)$ \\
\textit{inception\_4c-output}&512&$0.879/0.880 (\pm0.005)$&$0.869/0.868 (\pm0.005)$ \\
\textit{inception\_4d-output}&528&$0.880/0.880 (\pm0.006)$&$0.869/0.868 (\pm0.005)$ \\
\textit{inception\_4e-output}&832&$0.877/0.877 (\pm0.005)$&$0.867/0.867 (\pm0.007)$ \\
\textit{inception\_5a-output}&832&$0.878/0.878 (\pm0.007)$&$0.864/0.864 (\pm0.007)$ \\
\textit{inception\_5b-output}&1024&$0.865/0.865 (\pm0.007)$&$0.856/0.856 (\pm0.008)$ \\
 \hline
\textit{All concatenated}&5488&$\textbf{0.894/0.894} (\pm0.006)$&$\textbf{0.884/0.884} (\pm0.006)$\\ 
\hline
 \end{tabular}
\end{center}
\label{table:param1b}
\end{table*}

\begin{table*}[ht]
\caption{Performance comparison of deep features extracted from
Inception-V3's Inception modules without
fine-tuning measured on KonIQ-10k. 
Average/median $(\pm std)$ values are
reported over 100 random train-test splits. The best results are typed
by \textbf{bold}.
} 
\centering 
\begin{center}
    \begin{tabular}{ |c|c|c|c|}
    \hline
Layer&Dimension&PLCC&SROCC\\
    \hline
\textit{mixed0}&256&$0.843/0.843 (\pm0.006)$&$0.839/0.839 (\pm0.006)$\\
\textit{mixed1}&288&$0.848/0.848 (\pm0.005)$&$0.844/0.844 (\pm0.005)$\\
\textit{mixed2}&288&$0.849/0.849 (\pm0.006)$&$0.844/0.844 (\pm0.007)$\\
\textit{mixed3}&768&$0.861/0.860 (\pm0.005)$&$0.858/0.855 (\pm0.006)$\\
\textit{mixed4}&768&$0.897/0.897 (\pm0.005)$&$0.889/0.889 (\pm0.006)$\\
\textit{mixed5}&768&$0.906/0.906 (\pm0.004)$&$0.898/0.898 (\pm0.005)$\\
\textit{mixed6}&768&$0.902/0.901 (\pm0.004)$&$0.890/0.891 (\pm0.006)$\\
\textit{mixed7}&768&$0.884/0.884 (\pm0.004)$&$0.870/0.870 (\pm0.006)$\\
\textit{mixed8}&1280&$0.892/0.891 (\pm0.004)$&$0.879/0.879 (\pm0.006)$\\
\textit{mixed9}&2048&$0.871/0.871 (\pm0.005)$&$0.859/0.859 (\pm0.006)$\\
\textit{mixed10}&2048&$0.842/0.844 (\pm0.006)$&$0.828/0.829 (\pm0.008)$\\
 \hline
\textit{All concatenated}&10048&$\textbf{0.910/0.911} (\pm0.005)$&$\textbf{0.901/0.901} (\pm0.005)$\\ 
\hline
 \end{tabular}
\end{center}
\label{table:param2a}
\end{table*}

\begin{table*}[ht]
\caption{Performance comparison of deep features extracted from
Inception-V3's Inception modules with
fine-tuning measured on KonIQ-10k. 
Average/median $(\pm std)$ values are
reported over 100 random train-validation-test
splits. The best results are typed
by \textbf{bold}.
} 
\centering 
\begin{center}
    \begin{tabular}{ |c|c|c|c|}
    \hline
Layer&Dimension&PLCC&SROCC\\
    \hline
\textit{mixed0}&256&$0.848/0.848 (\pm0.008)$&$0.848/0.848 (\pm0.007)$\\
\textit{mixed1}&288&$0.853/0.853 (\pm0.007)$&$0.853/0.853 (\pm0.006)$\\
\textit{mixed2}&288&$0.854/0.853 (\pm0.007)$&$0.853/0.853 (\pm0.006)$\\
\textit{mixed3}&768&$0.866/0.865 (\pm0.006)$&$0.867/0.867 (\pm0.007)$\\
\textit{mixed4}&768&$0.902/0.902 (\pm0.007)$&$0.898/0.897 (\pm0.006)$\\
\textit{mixed5}&768&$0.911/0.910 (\pm0.005)$&$0.908/0.908 (\pm0.006)$\\
\textit{mixed6}&768&$0.907/0.906 (\pm0.005)$&$0.900/0.900 (\pm0.006)$\\
\textit{mixed7}&768&$0.889/0.889 (\pm0.005)$&$0.880/0.880 (\pm0.006)$\\
\textit{mixed8}&1280&$0.897/0.897 (\pm0.006)$&$0.888/0.887 (\pm0.008)$\\
\textit{mixed9}&2048&$0.876/0.876 (\pm0.005)$&$0.869/0.870 (\pm0.007)$\\
\textit{mixed10}&2048&$0.847/0.847 (\pm0.005)$&$0.837/0.836 (\pm0.008)$\\
 \hline
\textit{All concatenated}&10048&$\textbf{0.915/0.914} (\pm0.005)$&$\textbf{0.911/0.911} (\pm0.005)$\\ 
\hline
 \end{tabular}
\end{center}
\label{table:param2b}
\end{table*}

\begin{table*}[ht]
\caption{Publicly available IQA databases used in this study.
Publicly available IQA databases can be divided into two groups.
The first one contains a smaller set of reference images and 
artificially distorted images are derived from them using 
different noise types at
different intensity levels,
while the second one contains images with "natural" degradation without any
additional artificial distortions.  
} 
\centering 
\begin{center}
    \begin{tabular}{ |c|c|c|c|c|c|c|}
    \hline
\footnotesize{Database}&\footnotesize{Year}&\footnotesize{Reference images}&\footnotesize{Test images}&\footnotesize{Distortion type}& \footnotesize{Resolution} &\footnotesize{Subjective score}\\
    \hline
\footnotesize{LIVE In the Wild}&\footnotesize{2015}&\footnotesize{-}&\footnotesize{1,162}&\footnotesize{authentic}&\footnotesize{$500\times500$}&\footnotesize{MOS (1-5)} \\
\footnotesize{KonIQ-10k} & \footnotesize{2018} & \footnotesize{-}  &\footnotesize{10,073}& \footnotesize{authentic}  & \footnotesize{$1024\times768$}&\footnotesize{MOS (1-5)}  \\
\footnotesize{KADID-10k} & \footnotesize{2019} & \footnotesize{81} &\footnotesize{10,125}& \footnotesize{artificial} & \footnotesize{$512\times384$}&\footnotesize{MOS (1-5)}\\
 \hline
 \end{tabular}
\end{center}
\label{table:iqadatabase}
\end{table*}

\subsection{Comparison to the state-of-the-art}
\label{sec:comp}
To compare our proposed method to other state-of-the-art algorithms,
we collected \textit{ten} traditional learning-based
NR-IQA metrics (
DIIVINE \cite{moorthy2011blind},
BLIINDS-II \cite{saad2012blind},
BRISQUE \cite{mittal2012no},
CurveletQA \cite{liu2014no},
SSEQ \cite{liu2014noS},
GRAD-LOG-CP \cite{xue2014blind},
BMPRI \cite{min2018blind},
SPF-IQA \cite{varga2020no},
SCORER \cite{oszust2019local},
ENIQA \cite{chen2019no}
),
and \textit{two} opinion-unaware method (NIQE \cite{mittal2012making}, PIQE \cite{venkatanath2015blind})
whose original source code are available. Moreover, we took the results of \textit{two} recently
published deep learning based NR-IQA algorithms --- DeepFL-IQA \cite{lin2020deepfl} and
MLSP \cite{hosu2019effective} --- from their original
publication.
On the whole, we compared our proposed method
--- \textit{MultiGAP-NRIQA} --- to 12 other state-of-the-art
IQA algorithms or metrics.
The results can be seen in Table \ref{tab1}.

To ensure a fair comparison, these
traditional and deep methods were trained, tested, and evaluated
exactly the same as our proposed method. Specifically, $\sim 60\%$ of the
images was used for training, $\sim 20\%$ for validation, and $\sim 20\%$
for testing purposes. If a validation set is not required, the training
set contains $\sim 80\%$ of the images.
Moreover, for IQA databases containing artificial distortions the splitting
of the database is carried out with respect to the reference images, so
no semantic overlapping was between the training, validation, and test sets.
To compare our method to the state-of-the-art, we report on the
average PLCC and SROCC values
of 100 random train-validation-test splits of our method and those
of other algorithms.
As already mentioned, the results are summarized in
Table \ref{tab1}.
More specifically, this table illustrates
the measured average PLCC and SROCC
on three large publicly available IQA databases (Table \ref{table:iqadatabase} summarizes
the major parameters of the IQA databases used in this paper).

From the results, it can be seen that the proposed significantly outperforms
the state-of-the-art on \textbf{KonIQ-10k} database. Moreover, only the MultiGAP-NRIQA method is able
perform over 0.9 PLCC and SROCC. It can be observed that GPR with rational quadratic kernel
function performs better than SVR with Gaussian kernel function.
Similarly, the proposed method outperforms the state-of-the-art
on \textbf{LIVE In the Wild} IQA database \cite{ghadiyaram2015massive}
by a large margin.
On \textbf{KADID-10k}, DeepFL-IQA \cite{lin2020deepfl} provides the best results by a large margin. The proposed
MultiGAP-GPR gives the third best results.

\begin{table*}[ht]
\centering
\caption{Comparison of \textit{MultiGAP-NRIQA} with state-of-the-art NR-IQA
and FR-IQA algorithms trained and tested
on KonIQ-10k,
KADID-10k,
and TID2013 databases.
The average PLCC and SROCC values are reported measured over
100 random train-validation-test split.
The best results are shown in \textbf{bold} and the second best results are
typed in \textit{italic}.
The results of DeepFL-IQA \cite{lin2020deepfl} and
MLSP \cite{hosu2019effective} was measured by the authors of \cite{lin2020deepfl}.
}
\label{tab1}
\begin{tabular}{|c|c|c|c|c|c|c|}
\hline
&\multicolumn{2}{c|}{KonIQ-10k}&\multicolumn{2}{c|}{KADID-10k}&\multicolumn{2}{c|}{LIVE In the Wild}\\
Method &  PLCC & SROCC &  PLCC & SROCC &  PLCC & SROCC\\
\hline
DIIVINE \cite{moorthy2011blind} & 0.709& 0.692  &0.423 &0.428 & 0.602& 0.579 \\
BLIINDS-II \cite{saad2012blind} & 0.571 & 0.575 &0.548 &0.530 & 0.450& 0.419 \\
BRISQUE \cite{mittal2012no}   & 0.702& 0.676 & 0.383& 0.386& 0.503& 0.487\\
NIQE \cite{mittal2012making}  & - & -&0.273 &0.309 &- & -\\
CurveletQA \cite{liu2014no} & 0.728& 0.716 & 0.473& 0.450& 0.620& 0.611\\
SSEQ \cite{liu2014noS}      & 0.584& 0.573 &0.453 &0.433 & 0.469& 0.429 \\
GRAD-LOG-CP \cite{xue2014blind} & 0.705& 0.698 &0.585 &0.566& 0.579& 0.557 \\
PIQE \cite{venkatanath2015blind}    & 0.206& 0.245  &0.289 &0.237& 0.171& 0.108  \\
BMPRI \cite{min2018blind} & 0.636& 0.619 & 0.554& 0.530& 0.521& 0.480\\
SPF-IQA \cite{varga2020no} & 0.759& 0.740 & 0.717& 0.708& 0.592& 0.563\\
SCORER \cite{oszust2019local}  & 0.772& 0.762  & \textit{0.855}& \textit{0.856}& 0.599& 0.590 \\
ENIQA \cite{chen2019no} & 0.758& 0.744 & 0.634& 0.636& 0.578& 0.554\\
\hline
DeepFL-IQA \cite{lin2020deepfl} &  0.887&0.877 & \textbf{0.938} & \textbf{0.936} & \textit{0.842} & \textit{0.814}\\
MLSP \cite{hosu2019effective} & \textit{0.924} &\textit{0.913} & - & - &0.769 &0.734\\
\hline
MultiGAP-SVR \cite{varga2020multi} & 0.915& 0.911 &0.799 &0.795 &0.841& 0.813 \\
MultiGAP-GPR \cite{varga2020multi} & \textbf{0.928}&\textbf{0.925} &0.820 &0.814 & \textbf{0.857}&\textbf{0.826}\\

\hline
\end{tabular}
\end{table*}

\subsection{Cross database test}
To prove the generalization capability of our proposed MultiGAP-NRIQA method, we
carry out a so-called cross database test in this subsection. This means that our
model was trained on the whole KonIQ-10k \cite{lin2018koniq} database and tested
on LIVE In the Wild Image Quality Challenge Database \cite{ghadiyaram2015massive}.
Moreover, the other learning-based NR-IQA methods were also tested this way. The
results are summarized in Table \ref{cross}.
From the results, it can be clearly seen that all learning-based methods
performed significantly poorer in the cross database test
than in the previous tests. It should be emphasized that our MultiGAP-NRIQA
method generalized better than the state-of-the-art traditional or deep learning
based algorithms even without fine-tuning.
The performance drop occurs owing to the fact that images are treated
slightly differently in each publicly available IQA database.
For example, in LIVE In The Wild \cite{ghadiyaram2015massive} database the images
were rescaled. In contrast, the images of KonIQ-10k \cite{lin2018koniq} were cropped
from their original counterparts. 

\begin{table*}[ht]
\centering
\caption{Cross database test. The learning-based NR-IQA methods
were trained on the
whole KonIQ-10k database and tested on LIVE In
the Wild database.
The measured PLCC and SROCC values are reported.
The best results are shown in \textbf{bold} and the second best results are
typed in \textit{italic}. The results of DeepFL-IQA \cite{lin2020deepfl} was measured
by the authors of \cite{lin2020deepfl}.
}
\label{cross}
\begin{tabular}{|c|c|c|}
\hline
&\multicolumn{2}{c|}{LIVE In The Wild}\\
Method &  PLCC & SROCC \\
\hline
BLIINDS-II \cite{saad2011dct} &0.439&0.401\\
BRISQUE \cite{mittal2012no}   &0.625&0.604\\
DIIVINE \cite{moorthy2011blind}   &0.598&0.592\\
SSEQ \cite{liu2014no}        &0.440 &0.412\\
\hline
DeepFL-IQA \cite{lin2020deepfl} & - & 0.704  \\
MLSP \cite{hosu2019effective} & - & -\\
\hline
\textit{MultiGAP-NRIQA, RBF SVR} (without fine-tuning) &0.841&0.812\\
\textit{MultiGAP-NRIQA, RBF SVR} &0.841&0.813\\
\textit{MultiGAP-NRIQA, r.q. GPR} (without fine-tuning)&\textit{0.856}&\textit{0.855}\\
\textit{MultiGAP-NRIQA, r.q. GPR}&\textbf{0.857}&\textbf{0.856}\\
\hline
\end{tabular}
\end{table*}

\section{Conclusion}
\label{sec:conc}
In this paper, we introduced a deep framework for
NR-IQA which constructs a feature space
relying on multi-level Inception features extracted
from pretrained CNNs via GAP
layers.
Unlike previous deep methods, the proposed
approach do not take patches from the
input image, but instead treat the image
as a whole and extract image resolution
independent features. As a result, the proposed approach can be easily
generalized to any input image size and CNN base architecture.
Unlike previous deep methods, we extract multi-level features from the CNN
to incorporate both mid-level and high-level
deep representations into the feature
vector.
Furthermore, we pointed out in a detailed parameter study that mid-level
features provide significantly more effective descriptors for NR-IQA.
Another important observation was that the feature
vector containing both mid-level
and high-level representations outperforms all feature vectors containing the
representation of one level.
We also carried out a comparison to other
state-of-the-art methods and our approach outperformed the state-of-the-art
on the largest available benchmark IQA databases. Moreover, the results were
also confirmed in a cross database test. There are many directions for future
research. Specifically, we would like to improve the fine-tuning process in
order to transfer quality-aware features more effectively into the base CNN.
Another direction of future research could be the generalization of the applied
feature extraction method to other CNN architectures, such as residual networks.







\bibliographystyle{unsrt}  
\bibliography{references}  






\end{document}